\documentclass{article}

\usepackage[nonatbib,final]{nips_2018}

\usepackage{placeins}
\usepackage[utf8]{inputenc} 
\usepackage[T1]{fontenc}    
\usepackage{hyperref}       
\usepackage{url}            
\usepackage{booktabs}       
\usepackage{amsfonts}       
\usepackage{nicefrac}       
\usepackage{microtype}      
\usepackage{graphicx}
\usepackage{subfig}

\usepackage{biblatex}
\usepackage{booktabs} 
\usepackage{lipsum}
\usepackage{amsfonts,amsmath,amssymb,amsthm,mathtools} 
\usepackage{wrapfig}  
\usepackage{mdwlist}  
\usepackage{sidecap}  
\usepackage{bbm}
\usepackage{bm,amsbsy} 
\usepackage{algorithm,algorithmic}

\usepackage{caption}

\newcommand\blfootnote[1]{%
  \begingroup
  \renewcommand\thefootnote{}\footnote{#1}%
  \addtocounter{footnote}{-1}%
  \endgroup
}

\newcommand{\trp}{{^\top}} 

\renewcommand{\eqref}[1]{eq.~\ref{eq:#1}}
\newcommand{\Nrm}{\mathcal{N}}

\newcommand{\vx}{\mathbf{x}}

\newcommand{\Dat}{\mathcal{D}}

\newcommand{\vf}{\mathbf{f}}
\newcommand{\vb}{\mathbf{b}}
\newcommand{\va}{\mathbf{a}}
\newcommand{\valpha}{\mathbf{\ensuremath{\bm{\alpha}}}}
\newcommand{\vbeta}{\mathbf{\ensuremath{\bm{\beta}}}}

\newcommand{\vone}{\mathbf{1}} 

\newcommand{\vz}{\mathbf{z}}
\newcommand{\vw}{\mathbf{w}}

\newcommand{\mW}{\mathbf{W}}
\newcommand{\mZ}{\mathbf{Z}}
\newcommand{\mC}{\mathbf{C}}

\newcommand{\vy}{\mathbf{y}}

\begin{document}

\title{DP-MAC: The Differentially Private Method of Auxiliary Coordinates for Deep Learning}

\author{Frederik Harder$^1$
\And Jonas Köhler$^2$ 
\And Max Welling$^3$
\And Mijung Park$^4$
}\blfootnote{1,2,4: Max Planck Institute for Intelligent Systems, 3: University of Amsterdam}
\maketitle

\begin{abstract}

Developing a differentially private deep learning algorithm is challenging, due to the difficulty in analyzing the {\it{sensitivity}} of objective functions that are typically used to train deep neural networks. 
Many existing methods resort to the stochastic gradient descent  algorithm and apply
a {\it{pre-defined}} sensitivity to the gradients for privatizing  weights. 
However, their slow convergence
typically yields a high cumulative privacy loss. 
Here, we take a different route by employing the {\it{method of auxiliary coordinates}}, which allows us to independently update the weights per layer by optimizing a {\it{per-layer}} objective function. 
This objective function can be well approximated by a low-order Taylor's expansion, in which sensitivity analysis becomes tractable.
We perturb the coefficients of the expansion for privacy, which we  optimize using more advanced optimization routines than SGD for faster convergence. 
We empirically show that our algorithm provides a decent trained model quality under a modest privacy budget.\footnote{We updated this current manuscript by fixing an implementation error, which was part of the implementation to produce the results we presented at the PPML2018 workshop. For detailed comments on what changes we made, see Appendix.}
\end{abstract}

\section{Introduction}

While providing outstanding performance, it has been shown that trained deep neural networks (DNNs) can expose sensitive information from the dataset they were trained on
 \cite{Carlini_et_al_2018, Song17, Shokri15, Fredrikson15}. 
%
In order to protect potentially sensitive training data, many existing methods adopt the notion of privacy, called {\it{differential privacy}} (DP) \cite{Dwork14}. 
Differentially private algorithms often comprise a noise injection step (e.g. during the training process), which is generally detrimental to performance and leads to a trade-off between privacy and utility.  
The amount of noise necessary for a desired level of privacy depends on the \textit{sensitivity} of an algorithm, a maximum difference in its output depending on whether or not a single individual participates in the data. 
In DNNs, the sensitivity of an objective function is often intractable to quantify, since data appears in the function in a nested and complex way. In addition, such models have thousands to millions of parameters, one needs to saveguard, and require many passes over the dataset in training. As a result, providing meaningful privacy guarantees while maintaining reasonable performance remains a challenging task for DNNs.

One existing approach to this problem, DP-SGD \cite{2016arXiv160700133A, papernot:private-training, mcmahan2017}, avoids complicated sensitivities, by applying a pre-defined sensitivity to the gradients, which are then perturbed with Gaussian noise before updating the weights to ensure DP. 
This work also introduces the {\it{moments accountant}} (MA) \cite{2016arXiv160700133A}, a useful method for computing cumulative privacy loss when training for many epochs (a formal introduction to this method is found in Appendix A).
In another line of recent work \cite{Obj_Pertur_Taylor_12, DP_DeepAutoEnc, PP_ConvDeepBeliefNet17}, DP training is achieved by approximating the nested objective function through Taylor approximation and perturbing each of the coefficients of the approximated loss before training.

In this paper, we combine the benefits of these two approaches.
We modify the algorithm called the {\it{method of auxiliary coordinates}} (MAC), which allows independent weight updates per layer, by framing the interaction between layers as a local communication problem via introducing auxiliary coordinates  \cite{MAC14}. 
This allows us to split the nested objective function into per-layer objective functions, which can be approximated by low-order Taylor's expansions.
In this case the sensitivity analysis of the coefficients becomes tractable. 

\section{DP-MAC}\label{sec:DPMAC}

\subsection{The Method of Auxiliary Coordinates}

Here we provide a short introduction on the MAC algorithm (see \cite{MAC14}  for details). Under a fully connected neural net with $K$ hidden layers, a typical mean squared error (MSE) objective is given by 
\small{\begin{align}
\label{eq:nested_obj}
E(\mW) = \tfrac{1}{2N} \sum_{n=1}^N ||\vy_n - \vf(\vx_n; \mW)||^2,
\end{align}}
where $\vf(\vx_n; \mW) = \vf_{K+1}(\ldots  \vf_2(\vf_1(\vx_n;\mW_1)\ldots); \mW_{K+1})$. We denote $\mW$ as a collection of weight matrices of $(K+1)$-layers, i.e., $\mW = \{\mW_k\}_{k=1}^{K+1}$, where the size of each weight matrix is given by $\mW_k \in \mathbb{R}^{D^k_{in} \times D^k_{out}}$. Each layer activation function is given by $\vf_k(\vx_n; \mW_k) = \vf_k(\mW_k\trp \vx_n)$, and $\vf_k$ could be any type of element-wise activation functions. 
In the MAC framework \cite{MAC14}, the objective function in \eqref{nested_obj} is expanded by adding auxiliary variables $\{\vz_n\}$ (one per datapoint) such that the optimization over many variables are decoupled:
\begin{equation}\label{eq:obj_orig}
E(\mW, \mZ;\mu) = E_o(\mW, \mZ) + \sum_{k=1}^K E_k(\mW, \mZ, \mu), 
\vspace{-0.1cm}
 \end{equation}
where the partial objective functions at the output layer and at the $k$-th layer are given by
 $E_o(\mW_{K+1}, \mZ_K)= \frac{1}{2N} \sum_{n=1}^N ||\vy_n - \vf_{K+1}(\vz_{K,n}; \mW_{K+1})||^2 , $ and $
E_k(\mW_k, \mZ_k, \mZ_{k-1}, \mu) =  \frac{\mu}{2N} \sum_{n=1}^N  ||\vz_{k,n} - \vf_{k}(\vz_{k-1,n}; \mW_{k})||^2. $
Alternating optimization of this objective function w.r.t.\ $\mW$ and $\mZ$ minimizes the objective function. In this paper,  we set $\mu=1$, as suggested in  \cite{MAC14}.
For obtaining differentially private estimates of $\mW$, it turns out we  need to privatize the $\mW$ update steps only, while we can keep the $\mZ$ update steps non-private, as has been studied in Expectation Maximization (EM) type algorithms before \cite{park17c, VIPS16}.

To make this process DP, we first approximate each objective function as 1st or 2nd-order polynomials in the weights. Then, we perturb each approximate objective function by adding noise to the coefficients and optimize it for estimating $\mW$. 
How much noise we need to add to these coefficients depends on the sensitivity of these coefficients as well as the privacy loss we allow in each training step. 
The final estimate $\mW_T$ depends on estimates $\mW_t$ and $\mZ_t$ for all $t<T$ and so we keep the privacy loss per iteration fixed and compute the cumulative loss using the moments accountant.

\subsection{DP-approximate to the per-layer objective function}

First, we consider approximating the per-layer objective function 
via the 2nd-order Taylor expansion
\small{
\begin{align}\label{eq:k_th_layer_obj}
E_k(\mW_k) &=  \tfrac{1}{2N} \sum_{n=1}^N  ||\vz_{k,n} - \vf_{k}(\vz_{k-1,n}; \mW_{k})||^2
\approx  \tfrac{1}{2N}\left[a_k + \sum_{h=1}^{D^k_{out}} \vw_{kh}\trp\vb_{kh}  + \sum_{h=1}^{D^k_{out}} \vw_{kh}\trp \mC_{kh} \vw_{kh}\right],  \nonumber
\end{align}
}
where $\vw_{kh} \in \mathbb{R}^{D^k_{in}}$ is the $h$-th column of the matrix $\mW_k$, and the derivation of each term $a_k, \vb_{kh}\in \mathbb{R}^{D^k_{in}}, \mC_{kh}\in \mathbb{R}^{D^k_{in} \times D^k_{in}}$ is given below. Here we choose to use the {\it{softplus}} function as an example activation function
for $f$, but any twice differentiable function is valid.
We introduce a new notation $T_n(\vw_{kh})$ 
\small{
\begin{align}
E_k(\mW_k) =  \tfrac{1}{2N} \sum_{n=1}^N \sum_{h=1}^{D^k_{out}} T_n(\vw_{kh})
\end{align}
}
where $ T_n(\vw_{kh}) = 
z_{kh,n}^2 - 2 z_{kh,n}f({\vw_{kh}\trp\vz_{k-1,n}}) + \{f({\vw_{kh}\trp\vz_{k-1,n}})\}^2$.
We then approximate $T_n(\vw_{kh})$ by the 2nd-order Taylor expansion evaluated at $\hat{\vw}_{kh}$. In the first optimization step, we approximate the loss function by the 2nd-order Taylor expansion evaluated at a randomly drawn $\hat{\vw}_{kh} \sim \Nrm(0,I)$. In the consecutive optimization step, we evaluate the loss function at the noised-up estimate $\hat{\vw}_{kh}$ obtained from the previous optimization step.
{\small{
\begin{align}
T_n(\vw_{kh}) \approx T_n(\hat{\vw}_{kh}) + (\vw_{kh}-\hat{\vw}_{kh})\trp \partial T_{nkh} 
+ \tfrac{1}{2}(\vw_{kh}-\hat{\vw}_{kh})\trp \partial^2 T_{nkh}  (\vw_{kh}-\hat{\vw}_{kh}), 
\end{align}
}\normalsize}
where the derivative expressions of $T_n(\vw_{kh})$  are given by 
{\small{
\begin{align}
\partial T_{nkh} = & [- 2 z_{kh,n}f'({\hat\vw_{kh}\trp\vz_{k-1,n}})
+ 2f({\hat\vw_{kh}\trp\vz_{k-1,n}})f'({\hat\vw_{kh}\trp\vz_{k-1,n}})]
\vz_{k-1,n},\nonumber \\ 
\partial^2 T_{nkh}  = & [- 2 z_{kh,n}f''({\hat\vw_{kh}\trp\vz_{k-1,n}})
+ 2 f({\hat\vw_{kh}\trp\vz_{k-1,n}}) f''({\hat\vw_{kh}\trp\vz_{k-1,n}})
+ 2 \{f'({\hat\vw_{kh}\trp\vz_{k-1,n}})\}^2 ] \vz_{k-1,n}\vz_{k-1,n}\trp. \nonumber
\end{align}
}\normalsize}
From this, we define the coefficients $a_k, \vb_{kh}, \mC_{kh}$ as:
$
a_k =  \sum_{n=1}^N  \sum_{h=1}^{D^k_{out}} [ T_n(\hat{\vw}_{kh}) - \hat{\vw}_{kh}\trp \partial T_{nkh} 
+ \tfrac{1}{2}\hat{\vw}_{kh}\trp \partial^2 T_{nkh}\hat{\vw}_{kh} ], 
\vb_{kh} = \sum_{n=1}^N \left[ \partial T_{nkh} -  \partial^2 T_{nkh} \hat{\vw}_{kh} \right], \; \mC_{kh}\; = \sum_{n=1}^N \tfrac{1}{2} \partial^2 T_{nkh}.  
$

Adding Gaussian noise to these coefficients for privacy modifies the objective function by 
\small{\begin{align} \label{eq:betweenlayer_perturbed_obj}
\tilde{E}_k(\vw_k)& \approx 
\tfrac{1}{2N}\left[\tilde{a}_k + \sum_{h=1}^{D^k_{out}} \vw_{kh}\trp \tilde{\vb}_{kh}  + \sum_{h=1}^{D^k_{out}} \vw_{kh}\trp \tilde{\mC}_{kh} \vw_{kh}\right],
 \end{align}}
where $\tilde{a}_k = a_k + \Nrm(0, (\Delta a_k)^2\sigma^2 ), 
\tilde{\vb}_k = \vb_{k} + \Nrm(0, (\Delta \vb_k)^2\sigma^2 \mathbf{I} ),
\tilde{\mC}_k = \mC_{k} + \Nrm(0, (\Delta \mC_k)^2\sigma^2 \mathbf{I} )$ and the amount of additive Gaussian noise depends on the sensitivity $(\Delta a_k, \Delta \vb_k, \Delta \mC_k)$ of each term.
When using a purely gradient-based optimization routine (e.g. Adam, unlike Conjugate Gradient), we don't have to perturb $a_k$ and in the case of first order approximation $\mC_{k}$ is omitted as well, leaving only $\vb_k$. 
This method is not limited to MSE objectives and in the classification task we use a binary cross-entropy objective analogously in the output layer.

Note that on the first $\mW$ step, unperturbed 1st and 2nd-order approximations provide the same gradient.
In this case, if we use vanilla SGD to optimize this first order approximation, this boils down to a variant of DP-SGD, which optimizes each layer objective function separately.

Analytic Sensitivities of the coefficients
are given in the appendix. 
Depending on the architecture of a neural network and dataset at hand, these analytic sensitivity bounds can often be loose, in which case we propose to take a more direct approach and bound the sensitivities directly by clipping the norms of the coefficients $\|\partial T_{nk}\|_F \leq \Theta_{\partial T}$ and $\|\partial^2 T_{nk}\|_F \leq \Theta_{\partial^2 T}$. Here, $\partial T_{nk}$ denotes the matrix of $\partial T_{nkh}$ vectors, which are used to compute $\Delta \vb_k$ and $\Delta \mC_k$. We found that this yields significantly lower bounds in practice and use these clipping thresholds, along with linear Taylor expansion for all experiments listed below.

\subsection{Calculation of cumulative privacy loss}

Using the moments accountant and the theorem for subsampled Gaussian mechanism given in \cite{2016arXiv160700133A} for composition requires caution, since the log moments of privacy loss are linearly growing, only if we draw fresh noise per new subsampled data. Up to this point our algorithm, unfortunately, draws many noises for many losses given a subsampled data.  
This is fixed by treating the vectorized coefficients of the perturbed layerwise objectives as a single vector quantity $[a_1, \vb_1, \mC_1, \cdots, a_K, \vb_K, \mC_K,]$ which is perturbed in one step. As previously done in \cite{mcmahan2017}, we scale down each objective function coefficient by its own sensitivity times the number of partitions $\sqrt{MK}$, where $m=3$ if $a_k, \vb_k, \mC_k$ are used to compute the loss (and $m=1$ if only $\vb_k$ is used). This sets the concatenated vector's sensitivity to $1$. Then, we add the standard Gaussian noise with standard deviation $\sigma$ to the vectors and scale up each perturbed quantities by their own sensitivity times $\sqrt{MK}$. Partitioning the vector in this way allows us to effectively consider the sensitivity and clipping bounds in each layer independently. Details are given in appendix I.
The DP-MAC algorithm is summarized in Algorithm \ref{algo:DPDL-MAC}.

\begin{algorithm}[h]
\caption{DP-MAC  algorithm}\label{algo:DPDL-MAC}
\begin{algorithmic}
\vspace{0.1cm}
\REQUIRE Dataset $\Dat$, total number of iterations $T$, privacy parameter $\sigma^2$, sampling rate $q$
\vspace{0.1cm}
\ENSURE $(\varepsilon, \delta)$-DP weights $\{\mW_{k}\}_{k=1}^{K+1}$\\
\FOR{number of Iterations $\leq T$}
\STATE \textbf{1.} Optimize \eqref{obj_orig} for $\mZ$
\STATE \textbf{2.} Optimize \eqref{betweenlayer_perturbed_obj} (noised-up objective) for $\mW$
\ENDFOR
\STATE Calculate the total privacy loss $(\varepsilon, \delta)$ using moments accountant
\end{algorithmic}
\end{algorithm}

\section{Experiments}\label{sec:Experiments}

\paragraph{Autoencoder}
We examine the performance of DP-MAC, when training deeper models, in a reconstruction task with a fully connected autoencoder with 6 layers, as used in the original MAC paper \cite{MAC14}. Unlike the original paper, we don't store any $\vz$ values but initialized them with a forward pass on each iteration. For this, we use the USPS dataset is a collection of 16x16 pixel grayscale handwritten digits, of which we use 5000 samples for training and 5000 for testing. 
We provide results for $\varepsilon$ values of $1, 2, 4, 8$ with $\delta=$1e-5. 
For comparison, we train the same model with DP-SGD where we use a single norm bound $\Theta_g$ for the full step-wise gradient of the model.
In Fig 1, we observe that DP-MAC lacks behind DP-SGD in both private and non-private settings. We suspect that this is in part owed to vanishing gradient updates in the MAC model. We found that independent of the number of $Z$ updates per iteration, gradient updates in the first and last layer of the model differ by up to 4 orders of magnitude. DP-SGD does not exhibit this problem.

\begin{figure}[t]
    \centering
    \subfloat{
    \includegraphics[width=0.45\linewidth]{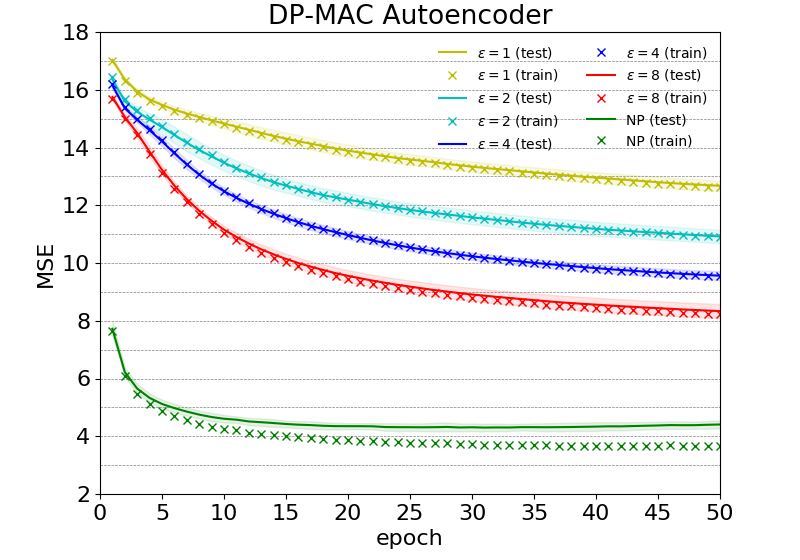}
    }
    \subfloat{
    \includegraphics[width=0.45\linewidth]{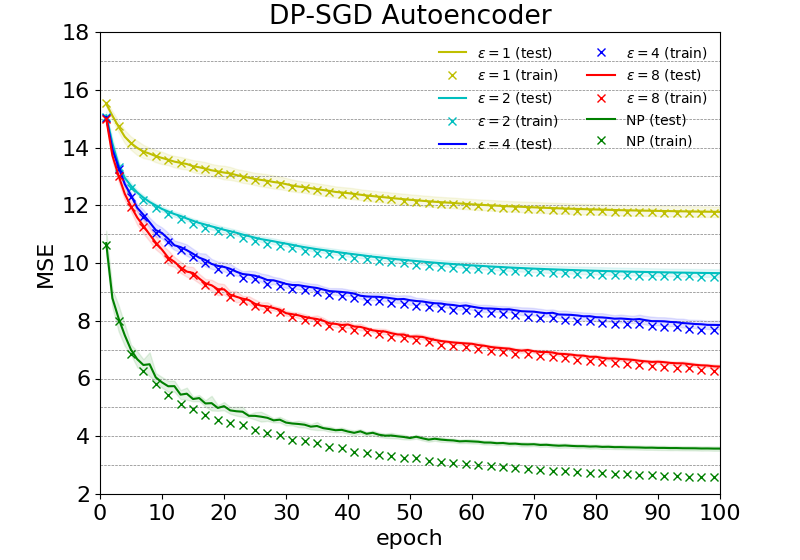}
    }
    \caption{Autoencoder training and test errors. ($\pm$ 1 stdev. of the latter) averaged over 10 runs each.}
    \label{fig:dp-ae-results}
\end{figure}

\paragraph{Classifier}
In addition, we show the performance of our method compared to other existing methods on a classification task on the MNIST digit dataset. We train a classifier with a single hidden layer of 300 units using DP-MAC with noise levels $\sigma=[1.0, 2.8, 8.0]$, which guarantees DP with $\delta=10^{-5}$ and $\varepsilon = [8, 2, 0.5]$. As in \cite{2016arXiv160700133A}, we use a DP-PCA to reduce input dimensionality to 60. Table \ref{fig:mnist-accuracy} shows the comparison between our method and the DP-SGD results by 
\cite{2016arXiv160700133A} as well as the DP convolutional deep belief networks (DP-CDBN) by \cite{PP_ConvDeepBeliefNet17}. Here, our method achieves a comparable test accuracy under the same privacy constraint within a relatively small number of training epochs.
Our implementation of both experiments is available at \mbox{\url{https://github.com/mijungi/dp_mac}}.

\begin{figure}[ht]
\centering
\begin{tabular}{|l||c|c|c|}
\hline
   & DP-SGD & DP-CDBN & \bf{DP-MAC} \\ 
\hline
$\varepsilon=0.5$ & $0.90$ &  $0.92$ & $0.90$ \\ 
\hline
$\#$ epochs & 16 &   162 & 10 \\ 
\hline
\hline
$\varepsilon=2$ & $0.95$ &  $0.95$ & $0.95$ \\ 
\hline
$\#$ epochs & 120 &   162 & 30 \\ 
\hline
\hline
$\varepsilon=8$ & $0.97$ &  & $0.97$ \\ 
\hline
$\#$ epochs & 700 & & 30 \\ 
\hline
\end{tabular}    
\caption{Test performance of DP-MAC compared to \cite{2016arXiv160700133A} DP-SGD and DP-CDBN \cite{PP_ConvDeepBeliefNet17} at  $\delta = 10^{-5}$}
\label{fig:mnist-accuracy}
\end{figure}

\section{Conclusion}
We present a novel differentially private deep learning paradigm, DP-MAC, which allows us to compute the sensitivity of the approximate objective functions analytically. Empirically however, we find that directly setting clipping bounds yields significantly lower sensitivities, which leads us to gradient perturbation as a special case. We found that MAC in its current state exhibits vanishing gradient problems in scaling to deeper models, which we believe causes the decrease in test performance compared to regular DP-SGD when training deeper models.
Nonetheless, we believe that this work offers an interesting new perspective on the possibilities computing sensitivities in deep neural networks.

\newpage

\nocite{SarwateC13}
\printbibliography

\newpage

\section*{Appendix A: Differential Privacy Background}
Here we provide background information on the definition of algorithmic privacy and a composition method that we will use in our algorithm, as well as the general formulation of the MAC algorithm. 

\subsection*{Differential privacy}
Differential privacy (DP) is a formal definition of the privacy properties of data analysis algorithms 
$[5]$.
Given an algorithm $\mathcal{M}$ and neighbouring datasets $\Dat$, $\Dat'$ differing by a single entry. Here, we focus on the inclusion-exclusion\footnote{This is for using the moments accountant method when calculating the cumulative privacy loss.} case, i.e., the dataset $\Dat'$ is obtained by excluding one datapoint from the dataset $\Dat$. The \emph{privacy loss} random variable of an outcome $o$ is
$L^{(o)} = \log \frac{Pr(\mathcal{M}_{(\Dat)} = o)}{Pr(\mathcal{M}_{(\Dat')} = o)} \mbox{ .}$
%
The mechanism $\mathcal{M}$ is called $\varepsilon$-DP if and only if
$|L^{(o)}| \leq \varepsilon, \forall o$.
A weaker version of the above is ($\varepsilon, \delta$)-DP, if and only if
$|L^{(o)}| \leq \varepsilon$, with probability at least $1-\delta$.
What the definition states is that a single individual's participation in the data do not change the output probabilities by much, which limits the amount of information that the algorithm reveals about any one individual.


The most common form of designing differentially private algorithms is by adding noise to a quantity of interest, e.g., a deterministic function $h: \Dat \mapsto \mathbb{R}^p$ computed on sensitive data $\Dat$. See 
$[5]$ and $[17]$
for more forms of designing differentially-private
algorithms. For privatizing $h$, one could use the {\it{Gaussian mechanism}}
$[16]$
which adds noise to the function, where the noise is calibrated to $h$'s {\it{sensitivity}}, $S_h$, defined by the maximum difference in terms of L2-norm, $\|h(\Dat)-h(\Dat') \|_2$,
$\tilde{h}(\Dat) = h(\Dat) + \Nrm(0, S_h^2\sigma^2 \mathbf{I}_p),$
where $\Nrm(0,S_h^2\sigma^2 \mathbf{I}_p)$ means the Gaussian distribution with mean $0$ and covariance $S_h^2\sigma^2 \mathbf{I}_p$.
The perturbed function $\tilde{h}(\Dat) $ is $(\varepsilon, \delta)$-DP, where $\sigma \geq \sqrt{2\log(1.25/\delta)}/\varepsilon$. 
In this paper, we use the Gaussian mechanism to achieve differentially private network weights. %
Next, we describe how the cumulative privacy loss is calculated when we use the Gaussian mechanism repeatedly during training.  

\subsection*{The moments accountant }
In the moments accountant, a cumulative privacy loss is calculated by bounding the moments of $L^{(o)}$, where 
the $\lambda$-th moment is defined as the log of the moment generating function evaluated at $\lambda$ 
$[6]$:
$\alpha_{\mathcal{M}}(\lambda; \Dat, \Dat') = \log \mathbb{E}_{o \sim \mathcal{M}(\Dat)} \left[ e^{\lambda  L^{(o)}}\right].$
By taking the maximum over the neighbouring datasets, we obtain the worst case $\lambda$-th moment
$ \alpha_{\mathcal{M}}(\lambda) = \max_{\Dat, \Dat'}\alpha_{\mathcal{M}}(\lambda; \Dat, \Dat')$, where the form of $\alpha_{\mathcal{M}}(\lambda)$ is determined by the mechanism of choice. The moments accountant compute $\alpha_{\mathcal{M}}(\lambda) $ at each step. Due to the composability theorem which states that
 the $\lambda$-th moment 
 composes linearly (See the composability theorem: Theorem 2.1 in 
 $[6]$
 when independent noise is added in each step, we can simply sum each upper bound on $\alpha_{\mathcal{M}_j}$ to obtain an upper bound on the total $\lambda$-th moment after $T$ compositions, 
$\alpha_{\mathcal{M}}(\lambda) \leq  \sum_{j=1}^T \alpha_{\mathcal{M}_j}(\lambda).$ 
%
Once the moment bound is computed, we can convert the $\lambda$-th moment to the ($\varepsilon,\delta$)-DP, guarantee by, 
$\delta = \min_{\lambda} \exp \left[ \alpha_{\mathcal{M}}(\lambda) - \lambda \varepsilon \right]$, for any $\varepsilon>0$. See Appendix A in 
$[6]$
for the proof.

\newpage
\section*{Appendix B: Experiment Results}

\begin{table}[ht]
    \centering
    \subfloat[Test classification accuracy on MNIST]{
\small{
\begin{tabular}{|l||c|c|c|}
\hline
   & DP-SGD & DP-CDBN & \bf{DP-MAC} \\ 
\hline
$\varepsilon=0.5$ & $0.90$ &  $0.92$ & $0.90$ \\ 
\hline
$\#$ epochs & 16 &   162 & 10 \\ 
\hline
\hline
$\varepsilon=2$ & $0.95$ &  $0.95$ & $0.95$ \\ 
\hline
$\#$ epochs & 120 &   162 & 30 \\ 
\hline
\hline
$\varepsilon=8$ & $0.97$ &  & $0.97$ \\ 
\hline
$\#$ epochs & 700 & & 30 \\ 
\hline
\end{tabular}\label{tab:mnist-accuracy}
    }}
    ~
    \subfloat[Test reconstruction MSE on USPS]{
\small{
\begin{tabular}{|l||c|c|c|}
\hline
  &  DP-SGD & \bf{DP-MAC} \\ 
\hline
$\varepsilon=1$  & $11.8$ &  $12.7$ \\ 
\hline
$\varepsilon=2$ & $9.6$ &  $10.9$ \\ 
\hline
$\varepsilon=4$ & $7.9$ &  $9.6$ \\ 
\hline
$\varepsilon=8$  & $6.4$ &  $8.4$\\ 
\hline
$\varepsilon=\infty$  & $3.6$ &  $4.4$\\ 
\hline
\end{tabular}\label{tab:ae-mse}
    }}
    \caption{Test performance of DP-MAC compared to $[6]$ DP-SGD and DP-CDBN $[11]$ at  $\delta = 10^{-5}$}
\end{table}

\begin{table}[ht!]
\begin{center}
\begin{tabular}{|l||c|c|c|}
\hline
 & DP-MAC Classifier    &  DP-MAC Autoencoder & DP-SGD Autoencoder\\ 
\hline
layer-sizes &  300 & 300-100-20-100-300-256 & 300-100-20-100-300-256 \\ 
\hline
batch size & 1000 & 500 (250 if $\varepsilon\leq 2$) & 500 (250  if $\varepsilon\leq 2$) \\ 
\hline
train epochs & 30 (10  if $\varepsilon=0.5$) & 50 & 100 \\ 
\hline
optimizer & Adam & Adam & SGD \\ 
\hline
$\mathbf{W}$ learning rate & 0.01 (0.03  if $\varepsilon=0.5$) & 0.003 & 0.03 \\ 
\hline
$\mathbf{z}$ learning rate & 0.003 & 0.001 &  \\ 
\hline
$\mathbf{W}$ lr-decay & 0.95 (0.7  if $\varepsilon=0.5$) & 0.97 & 100 (50  if $\varepsilon\leq2$) \\ 
\hline
$\mathbf{z}$-steps & 30 & 30 &  \\ 
\hline
$\mathbf{W}$-steps & 1 & 1 &  \\ 
\hline
$\Theta_{\partial T}$ & 0.3 & 0.001 &  \\ 
\hline
$\Theta_g$ &  &  & 0.01 \\ 
\hline
$\sigma$ values & 1.0, 2.8, 8.0 & 1.8, 3.1, 4.1, 7.8 & 2.4, 4.3, 5.7, 11.0 \\ 
\hline
$\sigma_{DP-PCA}$ & 4.0, 8.0, 16.0 &  &  \\ 
\hline
\end{tabular}
\caption{Training parameters choices for both DP-MAC experiments and the DP-SGD autoencoder comparison}
\end{center}
\end{table}

\newpage

\section*{Appendix C: Differences from the Previous Version}

The previous version of this paper contained an error in the implementation which mistakenly lowered the necessary amount of noise for a given privacy guarantee and, as a result reported wrong test results. In this version we have corrected this error and made a number of additional changes which are listed below:

\begin{itemize}
    \item Gradient update
    \begin{itemize}
        \item Fixed faulty gradient computation, which had reduced effective noise by up to 99\% during training.
        \item Improved clipping sensitivity by clipping $\partial T$ rather than the norms of the coefficients $\vb_k, \mC_k$.
        \item Reduced analytic sensitivity by 50\% by excluding $\tfrac{1}{2S}$ term from coefficients and making better use of inclusion/exclusion DP. 
    \end{itemize}
    \item Experiments
    \begin{itemize}
        \item Removed histogram-based layer-wise clipping bound search, which had turned out to be costly in terms of the privacy budget and yield relatively little improvement. Instead all layers now use the same bound.
        \item Classifier experiment now uses DP-PCA to reduce input dimensionality as in [6]. overall results stay roughly the same.
        \item Autoencoder: Worse results than DP-SGD, likely due to vanishing gradient issues.
        \item Replaced softplus activations with ReLUs.
        \item Significantly increased batch sizes. 
    \end{itemize}
    \item Notation
    \begin{itemize}
        \item Denoted Clipping thresholds as $\Theta$ to avoid confusion with $T_{nkh}$ terms.
        \item Defined coefficients $a_k, \vb_k, \mC_k$ excluding $\tfrac{1}{2S}$ term due to changes in sensitivity analysis.
    \end{itemize}
\end{itemize}

\newpage

\section*{Appendix D: Additional Figures}

\begin{figure}[ht]
\centering
\includegraphics[width=0.45\linewidth]{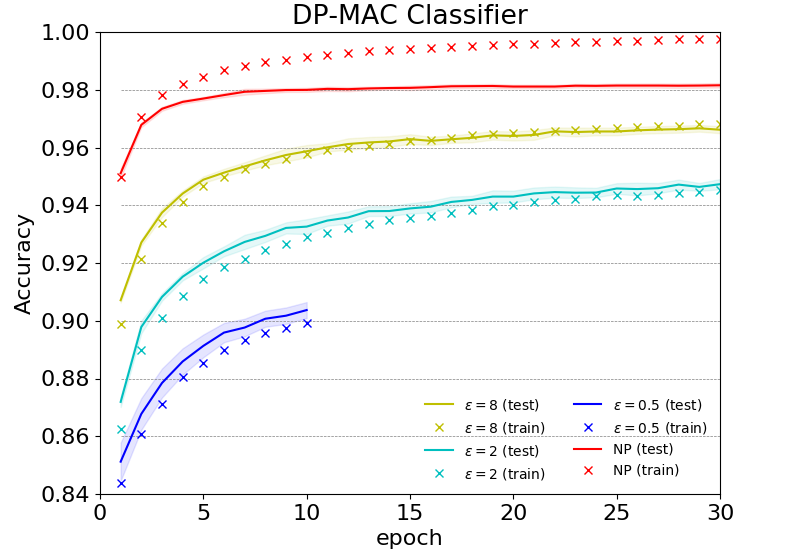}
\caption{Classifier training and test errors. ($\pm$ 1 stdev. of the latter) averaged over 10 runs each.}
\label{fig:dp-cl-results}
\end{figure}

\begin{figure}[ht]
        \centering
        \includegraphics[width=0.7\linewidth]
        {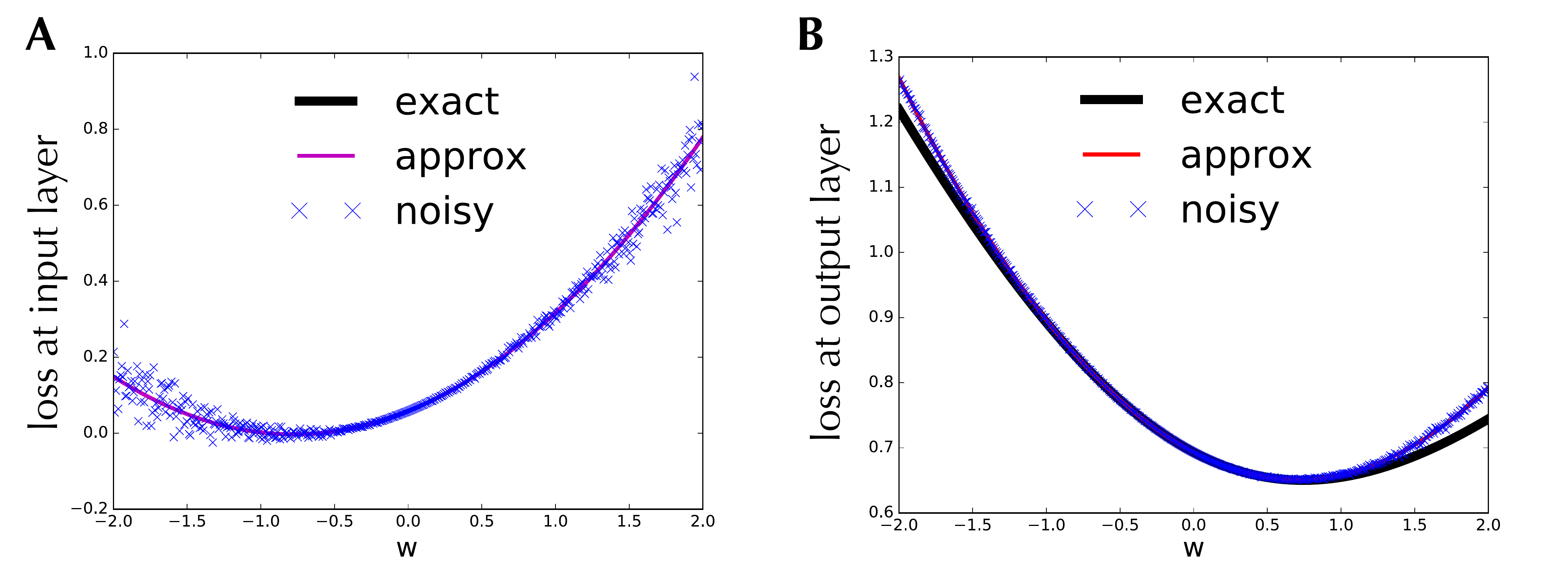}
        \caption{
 The input and output objective functions (black) are well approximated by the 2nd-order approximations (red). In both cases, approximation is made at 0, where the true $w$ at the input layer is $-0.7$, and $0.7$ at the output layer. The blue crosses depict additive noise centered around the approximated loss and the noise variance is determined by the sensitivities of the coefficients and privacy parameter $\sigma^2$.}
        \label{fig:comparison_approx}
\end{figure}

\begin{figure}[ht]
        \centering
        \includegraphics[width=0.45\linewidth]{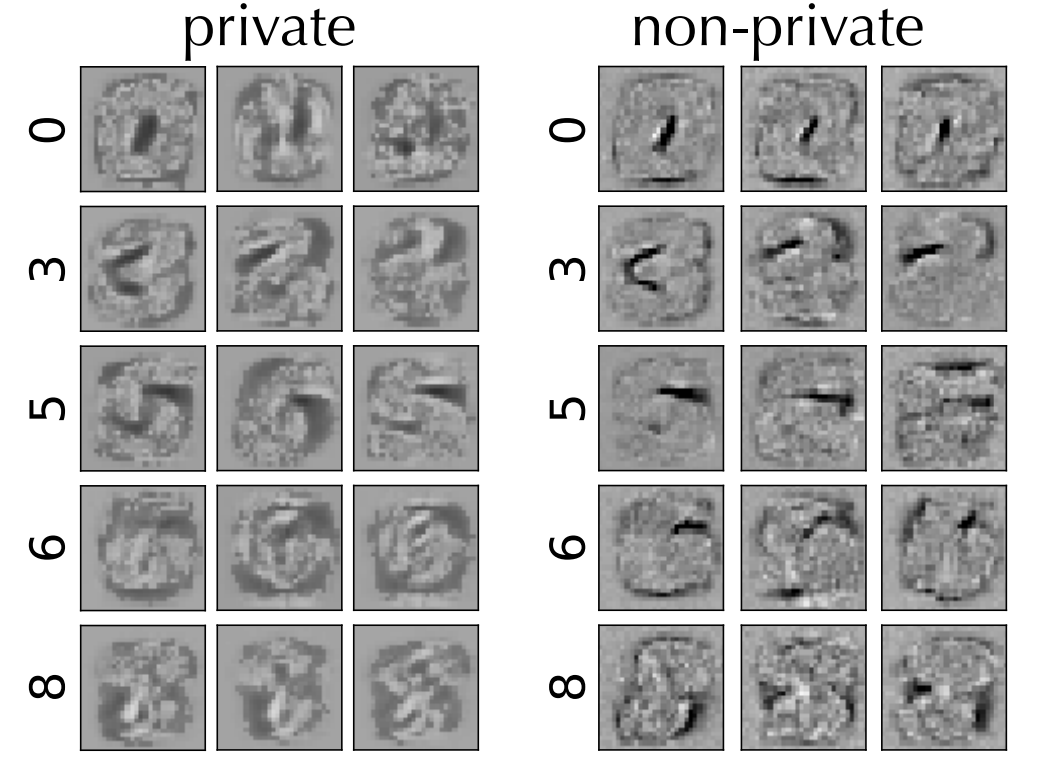}
        \caption{Learned significant features for labels 0,3,5,6,8 respectively. The non-private features show higher contrast and more characteristics in the high frequencies, whereas the private features become smoothed out and lose contrast.}
        \label{fig:dp-feature-diffs}
\end{figure}

\begin{algorithm}[ht]
\caption{DP-MAC with learning $T_{b_k}$ }\label{algo:DPDL-MAC-Tb}
\begin{algorithmic}
\vspace{0.1cm}
\REQUIRE $\Dat$, $T$, $\sigma^2$, $\sigma_{hist}^2$, $q$, initial threshold $T_{b_k}$
\vspace{0.1cm}
\ENSURE $(\varepsilon, \delta)$-DP weights $\{\mW_{k}\}_{k=1}^{K+1}$\\
\STATE \textbf{1.} Pre-training using DP-MAC (Algo. 1)
\STATE \textbf{2.} DP-histogram release which determines $T_{b_k}$
\STATE \textbf{3.} DP-MAC (Algo. 1) training using learned $T_{b_k}$
\end{algorithmic}
\end{algorithm}

\FloatBarrier

\newpage

\section*{Appendix E: sensitivity of $a_k$}

We are using a few assumptions and facts to derive sensitivities below. 
%
\begin{itemize}
\item $\|\vz_{k,s}\|_2 \leq T_z $ for a predefined threshold $T_z$ for all $k,s$.
\item Due to Cauchy-Schwarz inequality: $\vw_{kh}^T \vz_{k-1, s} \leq \|\vw_{kh}\|_2 T_z$
\item Using a monotonic nonlinearity (e.g., softplus): $f(\vw_{kh}^T \vz_{k-1, S}) \leq f(\|\vw_{kh}\|_2 T_z)$ and $f'(\vw_{kh}^T \vz_{k-1, S}) \leq f'(\|\vw_{kh}\|_2 T_z)$
\item For softplus, $ 0< f'(\vw_{kh}^T \vz_{k-1, S}) \leq f'(\|\vw_{kh}\|_2 T_z)\leq 1$ and $0< f^{''}(\vw_{kh} ^T \vz_{k-1, s}) \leq \frac{1}{4}$
\item $\|\va \|_2 \leq \| \va\|_1 \leq \|\va \|_2 \sqrt{D}$ for $\va \in \mathbb{R}^D$
\item Direct application of above : $|\sum_{h=1}^{D_{out}^k} z_{kh,s}| \leq  \sum_{h=1}^{D_{out}^k} |z_{kh,s}| = \|\vz_{k,s}|\trp \vone| \leq T_z \sqrt{D_{out}^k}$
\end{itemize}
%
%
Denote
$$\alpha_{\hat\vw_{kh}} := f(\|\hat\vw_{kh}\|_2 T_z) $$ 
$$\beta_{\hat\vw_{kh}} := f'(\|\hat\vw_{kh}\|_2 T_z) $$
which we will further denote as vectors $\valpha_{\hat\vw_k}, \vbeta_{\hat\vw_k}$.

Without loss of generality, we further assume that (1): the neighbouring datasets are in the form of $\Dat = \{\Dat', (\vx_S, \vy_S)\}$. 
\begin{align*}
&\Delta  a_k = \max_{|\Dat \setminus \Dat'|=1} |a_k(\Dat)- a_k(\Dat')|, \nonumber \\
&=\max_{|\Dat \setminus \Dat'|=1} | \sum_{h=1}^{D^k_{out}} \{ \sum_{s=1}^S (T_n(\hat{\vw}_{kh}) - \hat{\vw}_{kh}\trp\partial T_{skh} + \tfrac{1}{2}\hat{\vw}_{kh}\trp\partial^2 T_{skh}\hat{\vw}_{kh}) - \sum_{s=1}^{S-1} (T_n(\hat{\vw}_{kh}) - \hat{\vw}_{kh}\trp\partial T_{skh} + \tfrac{1}{2}\hat{\vw}_{kh}\trp\partial^2 T_{skh}\hat{\vw}_{kh})\}|\nonumber \\
&=\max_{|\Dat \setminus \Dat'|=1} | \sum_{h=1}^{D^k_{out}}
 (T_S(\hat{\vw}_{kh}) - \hat{\vw}_{kh}\trp\partial T_{Skh} + \tfrac{1}{2}\hat{\vw}_{kh}\trp\partial^2 T_{Skh}\hat{\vw}_{kh}) |
\nonumber
\end{align*} 
Now the sensitivity can be divided into three terms due to triangle inequality as
\begin{align*}
\Delta  a_k
& \leq  
\underbrace{\max_{|\Dat \setminus \Dat'|=1} \left| \sum_{h=1}^{D^k_{out}} T_S(\hat{\vw}_{kh}) \right|}_{\Delta a_{k_1}}
+ \underbrace{\max_{|\Dat \setminus \Dat'|=1} \left| \sum_{h=1}^{D^k_{out}} \hat{\vw}_{kh}\trp \partial T_{Skh}\right|}_{\Delta a_{k_2}} 
+ \underbrace{ \max_{|\Dat \setminus \Dat'|=1}  \left| \sum_{h=1}^{D^k_{out}}\tfrac{1}{2}\hat{\vw}_{kh}\trp \partial^2 T_{Skh}\hat{\vw}_{kh} \right|}_{\Delta a_{k_3}}. \nonumber
\end{align*}
We compute the sensitivity of each of these terms below. 
The sensitivity of $a_{k_1}$ is given by 
\begin{align*}
\Delta a_{k_1} &=  
\max_{\vz_{k,S},\vz_{k-1,S}} \left| \sum_{h=1}^{D^k_{out}} z_{kh,S}^2 - 2 z_{kh,S} f(\hat\vw^T_{kh} \vz_{k-1, S}) +  
\left(f(\hat\vw^T_{kh} \vz_{k-1, S})\right)^2 \right| \\
&\leq
\max_{\vz_{k,S},\vz_{k-1,S}} \left| \sum_{h=1}^{D^k_{out}} z_{kh,S}^2 \right| 
+ \left| \sum_{h=1}^{D^k_{out}} 2 z_{kh,S} f(\hat\vw^T_{kh} \vz_{k-1, S}) \right| 
+ \left| \sum_{h=1}^{D^k_{out}} \left(f(\hat\vw^T_{kh} \vz_{k-1, S})\right)^2 \right| \\
&\leq 
\left( T_z^2 + 2 T_z \|\valpha_{\hat\vw_k}\|_2 + 
\|\vbeta_{\hat\vw_k}\|_2^2 \right)
\end{align*}

The sensitivity of $a_{k_2}$ is given by 
\begin{align*}
\Delta a_{k_2} &= \max_{\vz_{k,S},\vz_{k-1,S}}  \left| \sum_{h=1}^{D^k_{out}} \left( 
- 2 z_{kh,S} f'(\hat\vw^T_{kh} \vz_{k-1, S}) 
+ 2 f(\hat\vw^T_{kh} \vz_{k-1, S}) f'(\hat\vw^T_{kh} \vz_{k-1, S}) \right) \hat\vw^T_{kh} \vz_{k-1, S} \right|\\
&\leq 
\max_{\vz_{k,S},\vz_{k-1,S}}  \left| \sum_{h=1}^{D^k_{out}} \left( 2 z_{kh,S} f'(\hat\vw^T_{kh} \vz_{k-1, S}) \right) \hat\vw^T_{kh} \vz_{k-1, S}  \right| 
+  \left| \sum_{h=1}^{D^k_{out}} \left( 2 f(\hat\vw^T_{kh} \vz_{k-1, S}) f'(\hat\vw^T_{kh} \vz_{k-1, S}) \right) \hat\vw^T_{kh} \vz_{k-1, S}  \right|\\
&\leq 
\max_{\vz_{k,S},\vz_{k-1,S}}  \left| \sum_{h=1}^{D^k_{out}} \| 2 z_{kh,S} f'(\hat\vw^T_{kh} \vz_{k-1, S})  \hat\vw_{kh}\|_2 \cdot \| \vz_{k-1, S} \|_2 \right| 
+ \left| \sum_{h=1}^{D^k_{out}} \| 2 f(\hat\vw^T_{kh} \vz_{k-1, S}) f'(\hat\vw^T_{kh} \vz_{k-1, S})  \hat\vw_{kh} \|_2 \cdot \| \vz_{k-1, S} \|_2 \right| \nonumber \\
&\leq 
2T_z \max_{\vz_{k,S},\vz_{k-1,S}} \left( \left| \sum_{h=1}^{D^k_{out}} |z_{kh,S}| \cdot \|  f'(\hat\vw^T_{kh} \vz_{k-1, S})  \hat\vw_{kh}\|_2 \right| 
+ \left| \sum_{h=1}^{D^k_{out}} \| f(\hat\vw^T_{kh} \vz_{k-1, S}) f'(\hat\vw^T_{kh} \vz_{k-1, S} )  \hat\vw_{kh} \|_2 \right| \right) \\
&\leq 
2T_z \left( 
T_z \cdot \left( \sum_{h=1}^{D^k_{out}} \beta^2_{\hat\vw_{kh}} \| \vw_{kh}\|^2_2  \right)^{1/2} 
+ \sum_{h=1}^{D^k_{out}} | \alpha_{\hat\vw_{kh}}  \beta_{\hat\vw_{kh}} | \cdot \|\hat\vw_{kh} \|_2 \right), 
\end{align*}

The sensitivity of $a_{k_3}$ is given by 

\begin{align*}
\Delta a_{k_3} &= 
\max_{\vz_{k,S},\vz_{k-1,S}} \left| \sum_{h=1}^{D^k_{out}} \left( 
- z_{kh,S} f''(\hat\vw^T_{kh} \vz_{k-1, S}) 
+ \left( f'(\hat\vw^T_{kh} \vz_{k-1, S}) \right)^2
+ f(\hat\vw^T_{kh} \vz_{k-1, S}) f''(\hat\vw^T_{kh} \vz_{k-1, S}) \right) \left(\hat\vw^T_{kh} \vz_{k-1, S} \right)^2 \right|\\
&\leq 
\max_{\vz_{k,S},\vz_{k-1,S}} \left| \sum_{h=1}^{D^k_{out}} \left( z_{kh,S} f''(\hat\vw^T_{kh} \vz_{k-1, S}) \right) \left( \hat\vw^T_{kh} \vz_{k-1, S} \right)^2 \right|
+ \left| \sum_{h=1}^{D^k_{out}} \left( \left( f'(\hat\vw^T_{kh} \vz_{k-1, S}) \right)^2 \right) \left( \hat\vw^T_{kh} \vz_{k-1, S} \right)^2\right| \\
&\qquad \qquad \quad + \left| \sum_{h=1}^{D^k_{out}} \left( f(\hat\vw^T_{kh} \vz_{k-1, S}) f''(\hat\vw^T_{kh} \vz_{k-1, S}) \right) \left( \hat\vw^T_{kh} \vz_{k-1, S} \right)^2 \right|\\
&\leq 
\max_{\vz_{k,S},\vz_{k-1,S}} \left| \sum_{h=1}^{D^k_{out}} z_{kh,S} \cdot f''(\hat\vw^T_{kh} \vz_{k-1, S}) \cdot  \|\hat\vw_{kh}\|_2^2 \cdot \|\vz_{k-1, S}\|_2^2 \right|
+ \left| \sum_{h=1}^{D^k_{out}} \left( f'(\hat\vw^T_{kh} \vz_{k-1, S}) \right)^2 \cdot \|\hat\vw_{kh}\|_2^2 \cdot \|\vz_{k-1, S}\|_2^2 \right| \\
&\qquad \qquad \quad + \left| \sum_{h=1}^{D^k_{out}} f(\hat\vw^T_{kh} \vz_{k-1, S}) f''(\hat\vw^T_{kh} \vz_{k-1, S}) \cdot \|\hat\vw_{kh}\|_2^2 \cdot \|\vz_{k-1, S}\|_2^2 \right|\\
&\leq
T_z^2 \max_{\vz_{k,S}} \left( \sum_{h=1}^{D^k_{out}} 1/4 \cdot  |z_{kh,S}| \cdot \|\hat\vw_{kh}\|_2^2 
+ \sum_{h=1}^{D^k_{out}} \left( \beta_{\hat\vw_{kh}} \right)^2 \cdot \|\hat\vw_{kh}\|_2^2 
+ \sum_{h=1}^{D^k_{out}} 1/4 \alpha_{\hat\vw_{kh}} \cdot \|\hat\vw_{kh}\|_2^2 \right)\\
&\leq 
\frac{T_z^2 }{4} \left( T_z \left( \sum_{h=1}^{D^k_{out}} \left( \|\vw_{kh}\|_2^2 \right)^2 \right)^{1/2}
+ \sum_{h=1}^{D^k_{out}} \left( 4 \left( \beta_{\hat\vw_{kh}} \right)^2
+ \alpha_{\hat\vw_{kh}} \right) \cdot \|\hat\vw_{kh}\|_2^2 \right)\\
\end{align*}

\section*{Appendix F: sensitivity of $\vb_k$}

\begin{align*}
\Delta  \vb_k &= \max_{|\Dat\setminus\Dat'|=1} \|\vb_{k}(\Dat)- \vb_{k}(\Dat')\|_F 
= \max_{|\Dat\setminus\Dat'|=1}  \left(\sum_{h=1}^{D_{out}^k}\|\vb_{kh}(\Dat)- \vb_{kh}(\Dat')\|^2_2\right)^{\tfrac{1}{2}}, \nonumber \\
&\leq 
\max_{\vz_{k,S} \vz_{k-1, S}, \vz'_{k,S}, \vz'_{k-1, S}}  \left(\sum_{h=1}^{D^k_{out}} \|(\partial T_S(\hat{\vw}_{kh}) 
- \partial^2 T_{S}(\hat{\vw}_{kh}) \hat{\vw}_{kh}) 
\|_2^2 \right)^{\tfrac{1}{2}}
\nonumber \\
&\leq 
\max_{\vz_{k,S} \vz_{k-1, S}, \vz'_{k,S}, \vz'_{k-1, S}} \left(\sum_{h=1}^{D^k_{out}} \|\partial T_{Skh}\|_2^2
+ \sum_{h=1}^{D^k_{out}} \|\partial^2 T_{Skh}\hat{\vw}_{kh}\|_2^2\right)^\frac{1}{2}
\\
&\leq 
(\Delta b_{k_1} + \Delta b_{k_2})^{\frac{1}{2}}, 
\end{align*}

\begin{align*}
\Delta  \vb_{k_1} &=   
\max_{\vz_{k,S},\vz_{k-1,S}} \sum_{h=1}^{D^k_{out}} \| \left(-2 z_{kh,S} f'(\hat\vw_{kh}^T \vz_{k-1,S})
+2 f(\hat\vw_{kh}^T \vz_{k-1,S}) f'(\hat\vw_{kh}^T \vz_{k-1,S})\right) \vz_{k-1,S} \|_2^2 \\
&\leq  
2 T_z^2 
\max_{\vz_{k,S} \vz_{k-1, S}}
\sum_{h=1}^{D^k_{out}} 
|(f(\hat\vw_{kh}^T \vz_{k-1,S}) - z_{kh,S})\beta_{\hat\vw_{kh}}|^2 
\\
&\leq 2 T_z^2
\max_{\vz_{k,S} \vz_{k-1, S}}
\sum_{h=1}^{D^k_{out}} 
|\alpha_{\hat\vw_{kh}}\beta_{\hat\vw_{kh}} - z_{kh,S}\beta_{\hat\vw_{kh}}|^2 
\\
&\leq 2 T_z^2
\max_{\vz_{k,S} \vz_{k-1, S}}
\sum_{h=1}^{D^k_{out}} 
\left(\alpha_{\hat\vw_{kh}}^2\beta_{\hat\vw_{kh}}^2 + 2\alpha_{\hat\vw_{kh}}\beta_{\hat\vw_{kh}}^2|z_{kh,S}|
+  z_{kh,S}^2\beta_{\hat\vw_{kh}}^2 \right)
\\
&\leq 2 T_z^2 \left(
||\valpha_{\hat\vw_k} \odot \vbeta_{\hat\vw_k}||^2_2 
+ 2\min\left(
T_z
\sum_{h=1}^{D^k_{out}} \alpha_{\hat\vw_{kh}}
,
\max_{\vz_{k,S} \vz_{k-1, S}}
\max_i\left(\alpha_{\hat\vw_{ki}}\beta_{\hat\vw_{ki}}^2\right)
\sum_{h=1}^{D^k_{out}} z_{kh,S}
\right)
 + T_z^2 \right), \mathrm{~since~z~may~be~negative}
\\
&\leq 2 T_z^2 \left(
||\valpha_{\hat\vw_k} \odot \vbeta_{\hat\vw_k}||^2_2
+ 2T_z\min\left(
\sum_{h=1}^{D^k_{out}} \alpha_{\hat\vw_{kh}}
,
\sqrt{D^k_{out}}
\max_i\left(\alpha_{\hat\vw_{ki}}\beta_{\hat\vw_{ki}}^2\right) \right)
 + T_z^2 \right)
\end{align*}

\begin{align*}
\Delta \vb_{k_2} &= 
\max_{\vz_{k,S},\vz_{k-1,S}}  \sum_{h=1}^{D^k_{out}} \| \Big( \Big( -2 z_{kh,S} f''(\hat\vw_{kh}^T \vz_{k-1,S})
+2 \left( f'(\hat\vw_{kh}^T \vz_{k-1,S}) \right)^2
\\&\qquad\qquad\qquad
+2 f(\hat\vw_{kh}^T \vz_{k-1,S}) f''(\hat\vw_{kh}^T \vz_{k-1,S}) \Big) \vz_{k-1,S} \vz_{k-1,S}^T \Big) \hat\vw_{kh} \|_2^2\\
&\leq 
2 \max_{\vz_{k,S},\vz_{k-1,S}}   \sum_{h=1}^{D^k_{out}} \| z_{kh,S} f''(\hat\vw_{kh}^T \vz_{k-1,S}) \vz_{k-1,S} \vz_{k-1,S}^T \hat\vw_{kh} \|^2_2
+ \| \left( f'(\hat\vw_{kh}^T \vz_{k-1,S}) \right)^2 \vz_{k-1,S} \vz_{k-1,S}^T \hat\vw_{kh}\|^2_2
\\&\qquad\qquad\qquad
+ \| f(\hat\vw_{kh}^T \vz_{k-1,S}) f''(\hat\vw_{kh}^T \vz_{k-1,S}) \vz_{k-1,S} \vz_{k-1,S}^T \hat\vw_{kh} \|^2_2\\
&\leq 
2 \max_{\vz_{k,S},\vz_{k-1,S}}   \sum_{h=1}^{D^k_{out}} \|  1/4 \cdot z_{kh,S} \vz_{k-1,S} \vz_{k-1,S}^T \hat\vw_{kh} \|^2_2
\\&\qquad\qquad\qquad
+ \| \left( \left( f'(\hat\vw_{kh}^T \vz_{k-1,S}) \right)^2 
+ 1/4 \cdot f(\hat\vw_{kh}^T \vz_{k-1,S}) \right) \vz_{k-1,S} \vz_{k-1,S}^T \hat\vw_{kh} \|^2_2\\
&\leq 
2 \max_{\vz_{k,S},\vz_{k-1,S}}  \sum_{h=1}^{D^k_{out}} |1/4 \cdot z_{kh,S}|^2 \cdot T_z^2 \cdot T_z^2 \|\hat\vw_{kh}\|^2_2
\\&\qquad\qquad\qquad
+ \| \left( \left( f'(\hat\vw_{kh}^T \vz_{k-1,S}) \right)^2 
+ 1/4 \cdot f(\hat\vw_{kh}^T \vz_{k-1,S}) \right)\hat\vw_{kh} \|^2_2 \cdot T_z^2 \cdot T_z^2  \\
&\leq 
\frac{T_z^4}{8} \left(T_z^2 \|\vw_{k}\|_{F}^2
+ \sum_{h=1}^{D^k_{out}} \|\left( 4 \beta^2_{\hat\vw_{kh}} 
+ \alpha_{\hat\vw_{kh}} \right) \hat\vw_{kh} \|^2_2 \right)\\
\end{align*}

\section*{Appendix G: sensitivity of $\mC_k$}

The sensitivity of $\Delta \mC_{k}$ is given by

\begin{align*}
\Delta  \mC_k &= \max_{|\Dat\setminus\Dat'|=1} \|\mC_k(\Dat)- \mC_k(\Dat')\|_F, \nonumber \\
&= \max_{|\Dat\setminus\Dat'|=1} \left(\sum_{h=1}^{D_{out}^k} \|\mC_{kh}(\Dat)- \mC_{kh}(\Dat')\|_F^2\right)^{\frac{1}{2}}
\end{align*}

where 

\begin{align*}
\mC_{kh}(\Dat) &=
\sum_{s=1}^S \tfrac{1}{2} \partial^2 T_{skh}, \nonumber \\
&=
\sum_{s=1}^S \left[- 2 z_{kh,s}f''({\hat\vw_{kh}\trp\vz_{k-1,s}})
+ 2 \{f'({\hat\vw_{kh}\trp\vz_{k-1,s}})\}^2 + 2 f({\hat\vw_{kh}\trp\vz_{k-1,s}}) f''({\hat\vw_{kh}\trp\vz_{k-1,s}})  \right] \vz_{k-1,s}\vz_{k-1,s}\trp. \nonumber
\end{align*}

Due to the triangle inequality, 
\begin{align*}
\Delta \mC_k &\leq
\max_{\vz_{k,S},\vz_{k-1,S}} \left( \sum_{h=1}^{D^k_{out}} \| \Delta \mC_{kh}\|_F^2 \right)^{1/2} \\
&= 
\max_{\vz_{k,S},\vz_{k-1,S}} \left( \sum_{h=1}^{D^k_{out}} \| \Big( -2 z_{kh,S} f''(\hat\vw_{kh}^T \vz_{k-1,S})
+2 \left( f'(\hat\vw_{kh}^T \vz_{k-1,S}) \right)^2
+2 f(\hat\vw_{kh}^T \vz_{k-1,S}) f''(\hat\vw_{kh}^T \vz_{k-1,S}) \Big)
\vz_{k-1,S} \vz_{k-1,S}^T \|_F^2 \right)^{1/2}\\
&\leq 
2T_z^2 \max_{\vz_{k,S},\vz_{k-1,S}} \left( \sum_{h=1}^{D^k_{out}} |z_{kh,S} f''(\hat\vw_{kh}^T \vz_{k-1,S})|^2
+ |\left( f'(\hat\vw_{kh}^T \vz_{k-1,S}) \right)^2
+ f(\hat\vw_{kh}^T \vz_{k-1,S}) f''(\hat\vw_{kh}^T \vz_{k-1,S})|^2 \right)^{1/2}\\
&\leq 
\frac{T_z^2 }{2} \left( T_z^2
+ \|4 \left( \vbeta_{\hat\vw_k} \right)^2
+ \valpha_{\hat\vw_k}\|_2^2 \right)^{1/2}
\end{align*}

\section*{Appendix H: sensitivity of coefficients in the output layer objective function}

\begin{align*}
\Delta a_o &\leq 
\frac{1}{2S} \max_{\vz_{K,S}} \left| \sum_{h=1}^{D^o_{out}} f(\hat\vw_{K+1h}^T \vz_{K,S}) 
- \vw_{{K+1}h}^T f'(\hat\vw_{K+1h}^T \vz_{K,S}) \vz_{K,S}
+ 1/2 \vw_{{K+1}h}^T f''(\hat\vw_{K+1h}^T \vz_{K,S}) \vz_{K,S} \vz_{K,S}^T \vw_{{K+1}h} \right| \\
&\leq 
\frac{1}{2S} \max_{\vz_{K,S}} \sum_{h=1}^{D^o_{out}} \left|  f(\hat\vw_{K+1h}^T \vz_{K,S}) \right|
+ \sum_{h=1}^{D^o_{out}} \left|  \vw_{{K+1}h}^T f'(\hat\vw_{K+1h}^T \vz_{K,S}) \vz_{K,S} \right|
+ \sum_{h=1}^{D^o_{out}} \left| 1/2 \vw_{{K+1}h}^T f''(\hat\vw_{K+1h}^T \vz_{K,S}) \vz_{K,S} \vz_{K,S}^T \vw_{{K+1}h} \right| \\
&\leq 
\frac{1}{2S} \left( \|\valpha_{\vw_{K+1}}\|_1 + T_z \sum_{h=1}^{D^o_{out}} \|\vw_{K+1h} \cdot \beta_{\vw_{K+1h}}\|_2 + \frac{T_z^2}{8} \|\vw_{K+1}\|^2_F \right)
\end{align*}

\begin{align*}
\Delta \vb_o &\leq
\frac{1}{2S} \max_{\vy, \vz_{K,S}} \left( \sum_{h=1}^{D^o_{out}} \| -y_h \vz_{K,S} 
+ f'(\hat\vw_{K+1h}^T \vz_{K,S}) \vz_{K,S} 
- f''(\hat\vw_{K+1h}^T \vz_{K,S}) \vz_{K,S} \vz_{K,S} ^T \vw_{{K+1}h} \|_2^2 \right)^{1/2} \\
&\leq 
\frac{1}{2S} \max_{\vy, \vz_{K,S}}  \left( \sum_{h=1}^{D^o_{out}} |(f'(\hat\vw_{K+1h}^T \vz_{K,S}) -y_h)
-f''(\hat\vw_{K+1h}^T \vz_{K,S}) \vz_{K,S} ^T \vw_{{K+1}h} |^2 \|\vz_{K,S} \|_2^2 \right)^{1/2} \\
&\leq 
\frac{T_z}{2S} \max_{\vy, \vz_{K,S}}  \Bigg( \sum_{h=1}^{D^o_{out}} (f'(\hat\vw_{K+1h}^T \vz_{K,S}) -y_h)^2 + 2|(f'(\hat\vw_{K+1h}^T \vz_{K,S}) -y_h)f''(\hat\vw_{K+1h}^T \vz_{K,S}) \vz_{K,S} ^T \vw_{{K+1}h}| 
\\
&\qquad\qquad\qquad\qquad + (f''(\hat\vw_{K+1h}^T \vz_{K,S}) \vz_{K,S} ^T \vw_{{K+1}h})^2 \Bigg)^{1/2} \\
&\leq 
\frac{T_z}{2S} \left( D^o_{out} 
+  2 T_z \sum_{h=1}^{D^o_{out}}[ \beta_{\vw_{K+1}h} \|\vw_{{K+1}h}\|_2] + 1/16 \cdot T_z^2 \cdot \|\mW_{K+1}\|^2_F \right)^{1/2}
\end{align*}

\begin{align*}
\Delta \mC_o &\leq 
\frac{1}{2S} \max_{\vz_{K,S}} \left(\sum_{h=1}^{D^o_{out}} \|1/2 f''(\hat\vw_{K+1h}^T \vz_{K,S}) \vz_{K,S} \vz_{K,S}^T \|_F^2 \right)^{1/2} \\
&\leq 
\frac{1}{2S} \max_{\vz_{K,S}} \left(\sum_{h=1}^{D^o_{out}} \| 1/8 \vz_{K,S} \vz_{K,S}^T \|_F^2 \right)^{1/2} \\
&\leq
\frac{1}{16S} \sqrt{D^o_{out}} T_z^2
\end{align*}


\section*{Appendix I: Computing a cumulative privacy loss}

\subsection*{Preliminary}

We first address how the level of perturbation in the coefficients affects the level of privacy in the resulting estimate. 
Suppose we have an objective function that's quadratic in $w$, i.e., 
\begin{align*}
E(w) = a + bw + cw^2, 
\end{align*} where only the coefficients $a,b,c$ contain the information on the data (not anything else in the objective function is relevant to data).
We perturb the coefficients to ensure the coefficients are collectively $(\epsilon, \delta)$-differentially private. 
\begin{align*}
\tilde a &= a + n_a, \mbox{ where } n_a \sim \Nrm(0, \Delta_a^2 \sigma^2),\\
\tilde b &= b +  n_b, \mbox{ where } n_b \sim \Nrm(0, \Delta_b^2 \sigma^2 ),\\
\tilde c & =  c  + n_c, \mbox{ where } n_c \sim  \Nrm(0, \Delta_c^2 \sigma^2 ),
\end{align*} where $\Delta_a, \Delta_b, \Delta_c$ are the sensitivities of each term, and $\sigma$ is a function of $\epsilon$ and $\delta$. 
Here ``collectively" means composing the perturbed $\tilde{a}, \tilde{b}, \tilde{c}$ results in ($\epsilon, \delta$)-DP. 
For instance, if one uses the linear composition method (privacy degrades with the number of compositions), and perturbs each of these with $\epsilon_a$, $\epsilon_b$, and $\epsilon_c$, then the total privacy loss should match the sum of these losses, i.e., $\epsilon = \epsilon_a + \epsilon_b + \epsilon_c$. In this case, if one allocates the same privacy budget to perturb each of these coefficients, then $\epsilon_a=\epsilon_b=\epsilon_c = \epsilon/3$. The same holds for $\delta$.  

However, if one uses more advanced composition methods and allocates the same privacy budget for each perturbation, per-perturbation budget becomes some function (denoted by $g$) of total privacy budget $\epsilon$, i.e., $\epsilon_a=\epsilon_b=\epsilon_c = g(\epsilon)$, where $g(\epsilon) \geq \epsilon/3$. So, per-perturbation for $a, b, c$ has a higher privacy budget to spend, resulting in adding less amount of noise. 

Whatever composition methods one uses to allocate the privacy budget in each perturbation of those coefficients, since the objective function is a simple quadratic form in $w$, the resulting estimate of $w$ is some function of those perturbed coefficients, i.e., $\hat{w} = h(\tilde{a}, \tilde{b}, \tilde{c})$. Since the data are summarized in the coefficients and the coefficients are ($\epsilon, \delta$)-differentially private, the function of these coefficients is also  ($\epsilon, \delta$)-differentially private. 

One could write the perturbed objective as 
\begin{align*}
\tilde{E}(w) &= \tilde{a} + \tilde{b}w + \tilde{c}w^2,\\
&= (a + bw + cw^2) + (n_a + n_b w + n_c w^2), \\
&= E(w) + n(w).
\end{align*} Note that we write down the noise term as $n(w)$ to emphasize that when we optimize this objective function, the noise term also contributes to the gradient with respect to $w$ (not just the term $E(w)$). 

If we denote some standard normal noise $\alpha \sim \Nrm(0,1)$, we can rewrite the noise term as 
\begin{align*}
    n(w) = (\Delta_a + \Delta_b w + \Delta_c w^2) \sigma \alpha,
\end{align*} which is equivalent to 
\begin{align*}
    n(w) & \sim \Nrm(0, (\Delta_a + \Delta_b w + \Delta_c w^2)^2 \sigma^2), \\
         & \sim \Nrm(0, \Delta_{E(w)}^2 \sigma^2)
\end{align*} where we denote $\Delta_{E(w)}= \Delta_a + \Delta_b w + \Delta_c w^2$.

\subsection*{Extending the preliminary to DP-MAC}

In the framework of DP-MAC, given a mini-batch of data $\Dat_q$ with a sampling rate $q$, the DP-mechanism we introduce first computes coefficients for layer-wise objective functions ($K$ layer-wise objective functions for a model with $K$ layers, including the output layer), then noise up the coefficients using Gaussian noise, and outputs the vector of perturbed coefficients for each layer, given by:
\begin{align*}
\mathcal{M}_k(\Dat_q) 
&= 
\begin{bmatrix}
a_k  \\[0.3em]
\vb_k  \\[0.3em]
\mC_k  \\[0.3em]
\end{bmatrix} + 
\begin{bmatrix}
n^*_{a,k}(\mW_k, \Delta_{a_k}) \\[0.3em]
n^*_{\vb,k}(\mW_k, \Delta_{\vb_k}) \\[0.3em]
n^*_{\mC,k}(\mW_k, \Delta_{\mC_k}) \\[0.3em]
\end{bmatrix}.
\end{align*} 
We denote the noise terms by $n^*(\mW, \Delta)$ and the sensitivities of each coefficient by $\Delta_{a_k}, \cdots, \Delta_{\mC_k}$.

Here the question is, if we decide to use an advanced composition method such as moments accountant, how the log-moment of the privacy loss random variable composes in this case. To directly use the composition theorem of Abadi et al, we need to draw a fresh noise whenever we have a new subsampled data. This means, there should be an instance of Gaussian mechanism that affects the these noise terms simultaneously.

To achieve this, we rewrite the vector of perturbed objective coefficients as $\tilde{\mathbf{E}}(\vw)$ below. For each layer we gather the loss coefficients into one vector $[a_k, \mathrm{vec}(\vb_k), \mathrm{vec}(\mC_k)]^T$. Then, we scale down each objective function by its own sensitivity times the number of partitions $\sqrt{MK}$, so that the concatenated vector's sensitivity becomes just $1$. Then, add the standard normal noise to the vectors with scaled standard deviation, $\sigma$. Then, scale up each perturbed quantities by its own sensitivity times $\sqrt{MK}$. In this example we use all three coefficients, so $M=3$. Note that in the experiments, using linear expansion we would only use $\vb_k$ and so $M$ would equal $1$ in that case. In the following we use $P_{m,k}$ to denote the a partition of the vector, which may pick out any of the contained coefficients, e.g. $a_k, \vb_k$ or $\mC_k$ for $m=1,m=2$ and $m=3$ respectively.
\begin{align*}
\mathcal{M}(\Dat_q)
&=
\begin{bmatrix}
\tilde{P}_{1,1}(\mW_1)  \\[0.3em]
\vdots \\[0.3em]
\tilde{P}_{M,1}(\mW_1) \\[0.3em]
\vdots \\[0.3em]
\tilde{P}_{M,K}(\mW_K) \\[0.3em]
\end{bmatrix} \\
&=
\begin{bmatrix}
P_{1,1}(\mW_1)  \\[0.3em]
\vdots \\[0.3em]
P_{M,K}(\mW_K) \\[0.3em]
\end{bmatrix} +
\begin{bmatrix}
n^*_1(\mW_1) \\[0.3em]
\vdots \\[0.3em]
n^*_{M,K}(\mW_K) \\[0.3em]
\end{bmatrix} \\
&=
\begin{bmatrix}
\sqrt{MK}\Delta_{P_{1,1}(\mW_1)} \cdot \left\{ \frac{P_{1,1}(\mW_1)}{\sqrt{MK}\Delta_{P_{1,1}(\mW_1)}} + \sigma \Nrm(0,1) \right\}  \\[0.3em]
\ldots  \\[0.3em]
\sqrt{MK}\Delta_{P_{M,K}(\mW_K)} \cdot \left\{ \frac{P_{M,K}(\mW_K)}{\sqrt{MK}\Delta_{P_{M,K}(\mW_K)}} + \sigma \Nrm(0,1) \right\}  \\[0.3em]
\end{bmatrix} \\
& =
\begin{bmatrix}
\sqrt{MK}\Delta_{P_{1,1}(\mW_1)}   \\[0.3em]
\ldots  \\[0.3em]
\sqrt{MK}\Delta_{P_{M,K}(\mW_K)}   \\[0.3em]
\end{bmatrix}
\cdot
\left(    \begin{bmatrix}
\frac{P_{1,1}(\mW_1)}{\sqrt{K}\Delta_{P_{1,1}(\mW_1)}}  \\[0.3em]
\ldots  \\[0.3em]
\frac{P_{M,K}(\mW_K)}{\sqrt{K}\Delta_{P_{M,K}(\mW_K)}}   \\[0.3em]
\end{bmatrix}
+
\Nrm(0, \sigma^2 I) 
\right)
\end{align*} 

Since we're adding independent Gaussian noise under each subsampled data, the privacy loss after $T$ steps, is simply following the composibility theorem in the Abadi et al paper. 

So compared to the sensitivity for $n(w)$ in the first section, the new noise $n^*(w)$ has a higher sensitivity due to the factor $\sqrt{MK}$.

\subsection*{Moments Calculations}

In this case, with a subsampling with rate $q$, we re-do the calculations in Abadi et al. First, let:
\begin{align*}
    \mu_0 = \mathcal{N}(\mathbf{0}_K, \sigma^2 I),
    \mu_1 = \mathcal{N}(\mathbf{1}_K, \sigma^2 I)
\end{align*} and let $\mu$ as a mixture of the two Gaussians, 
\begin{align*}
 \mu = (1 - q) \mathcal{N}(\mathbf{0}_K, \sigma^2 I) + q \mathcal{N}(\mathbf{1}_K, \sigma^2 I).
 \end{align*}
%
Here $\mathbf{0}_K$ is the $K$-dimensional $0$ vector, and $\mathbf{1}_K$ is the $K$-dimensional all ones vector. 
Here $\alpha_M(\lambda)$ should be $\log \max(E_1, E_2)$ where 
\[ E_1 = \mathbb{E}_{z \sim \mu}[(\mu(z)/\mu_0(z))^{\lambda}], E_2 = \mathbb{E}_{z \sim \mu_0}[(\mu_0(z)/\mu(z))^{\lambda}] \]
Then, we can compose further mechanisms using this particular $\alpha_M(\lambda)$, which follows the same analysis as in Abadi et al.

\end{document}